\documentclass[journal]{IEEEtran}

\usepackage{cite}
\ifCLASSINFOpdf
   \usepackage[pdftex]{graphicx}
\else
   \usepackage[dvips]{graphicx}
\fi
\usepackage{amsmath}
\usepackage{stfloats}
\fnbelowfloat
 \usepackage{float} 
\usepackage{url}
\usepackage[switch]{lineno}
\linenumbers
\usepackage{enumerate}
\usepackage[dvipsnames]{xcolor}
\usepackage[center]{caption}

\begin{document}
\nolinenumbers

\title{Identity masking effectiveness and gesture recognition: Effects of eye enhancement in seeing through the mask }

\author{\IEEEauthorblockN{Madeline Rachow\IEEEauthorrefmark{1},
Thomas Karnowski\IEEEauthorrefmark{2} and Alice~J. O'Toole\IEEEauthorrefmark{3}}\\
\IEEEauthorblockA{\IEEEauthorrefmark{1}University of Arkansas\\ 
\IEEEauthorrefmark{2} Oak Ridge National Laboratory\\
\IEEEauthorrefmark{1} The University of Texas at Dallas\\
Email: \IEEEauthorrefmark{1}mrachow@uark.edu,
\IEEEauthorrefmark{2}karnowskitp@ornl.gov,
\IEEEauthorrefmark{3}otoole@utdallas.edu
}
}


\markboth{Identity masking with eye enhancement}%
{Evaluating the Effectiveness of Automated Identity Masking}

\maketitle

\begin{abstract}

Face identity masking algorithms developed in recent years aim to protect the privacy of people in video recordings.  These algorithms are designed
 to interfere with identification, while
preserving information about facial actions.
An important challenge is to preserve subtle actions in the eye region, while obscuring
the salient identity cues from the eyes.
We evaluated the effectiveness of identity-masking 
algorithms based on Canny filters, applied
with and without eye enhancement,
for interfering with identification and preserving facial actions.  In Experiments 1 and 2, we tested human participants' ability to
match the facial identity of a driver in a low resolution video to a high resolution facial image. Results showed that both masking methods impaired identification, and that eye enhancement did not alter the effectiveness of the Canny filter mask.  In Experiment 3, we tested action preservation and found that neither method interfered significantly with driver action perception. We conclude that relatively simple,
filter-based masking algorithms, which are suitable for application to low quality video, can be used in privacy protection without compromising action perception.

\end{abstract}


\IEEEpeerreviewmaketitle


\section{Introduction}
Video recordings for security and surveillance are now ubiquitous in public and private spaces. This has lead to a pressing need to develop face identity masking algorithms aimed at protecting the privacy of people in the recordings.  Facial identity masking technology 
also needs to preserve the facial actions (gestures and expressions) of those being photographed for applications that require action classification without identification. 
Understanding and measuring the extent to which identity-masking algorithms effectively accomplish both goals is a challenging problem. Because
identification and action classification are tasks that can be done accurately by humans, the success of masking algorithms cannot be evaluated comprehensively 
without examining human perception.

Human identification and gesture categorization of identity-masked faces 
have been examined previously \cite{hooge2020evaluating}. The effectiveness of eight different identity masking algorithms was evaluated using  human perception and a deep convolutional neural network (DCNN) trained for face identification. 
Human participants and the DCNN were tested 
with videos taken of drivers actively operating a motor vehicle. For the human experiment, people studied high-resolution images of the drivers to learn their identities and were tested  on their recognition of those drivers in low-resolution 
 videos.  Test videos were low resolution and showed drivers actively operating a motor vehicle. Videos were either unmasked or  masked by one of eight algorithms, including methods that rely on Facial Action Transfer (FAT) (cf., \cite{Huang2012,xiong2013supervised}), a DMask \cite{DMask},  Canny filtering \cite{Canny1986}, and 
Scharr filtering \cite{Jahne1999}.
The results showed that all of the algorithms reduced  human face recognition accuracy.  Moreover, people made their recognition decisions with a conservative  response bias (i.e., a tendency to indicate that they did not recognize drivers, when they were uncertain). This bias  indicates that the participants had low confidence in their identification decisions---supporting the effectiveness of the masking methods. 

In the machine evaluation of that 
test \cite{hooge2020evaluating}, the DCNN matched identities between the high-resolution images and masked videos, and between the unmasked and masked videos. DCNN performance matching high-resolution images to masked and unmasked videos showed a pattern of poor performance approximately comparable to human behavior---echoing the effectiveness of the masking algorithms for both humans and the CNN. The results showed that even simple methods, such as edge-detection, can impair identification performance. 

It is worth noting that
 more sophisticated methods than filtering have been developed for identity masking, including  generative adversarial networks, GANS (e.g., \cite{Khojaste22}). However, these techniques can only be applied to high quality (frontal) images and are computationally intense, which limits their utility for high volume throughput 
 (e.g., videos). 
Many important applications of face identity masking must deal with large quantities of low resolution, poor quality video. Therefore, there is a need to consider the effectiveness of simpler methods that can be applied in these less controlled circumstances. 

   \begin{figure*}[h]
      \centering
      \includegraphics[width=.70\textwidth]{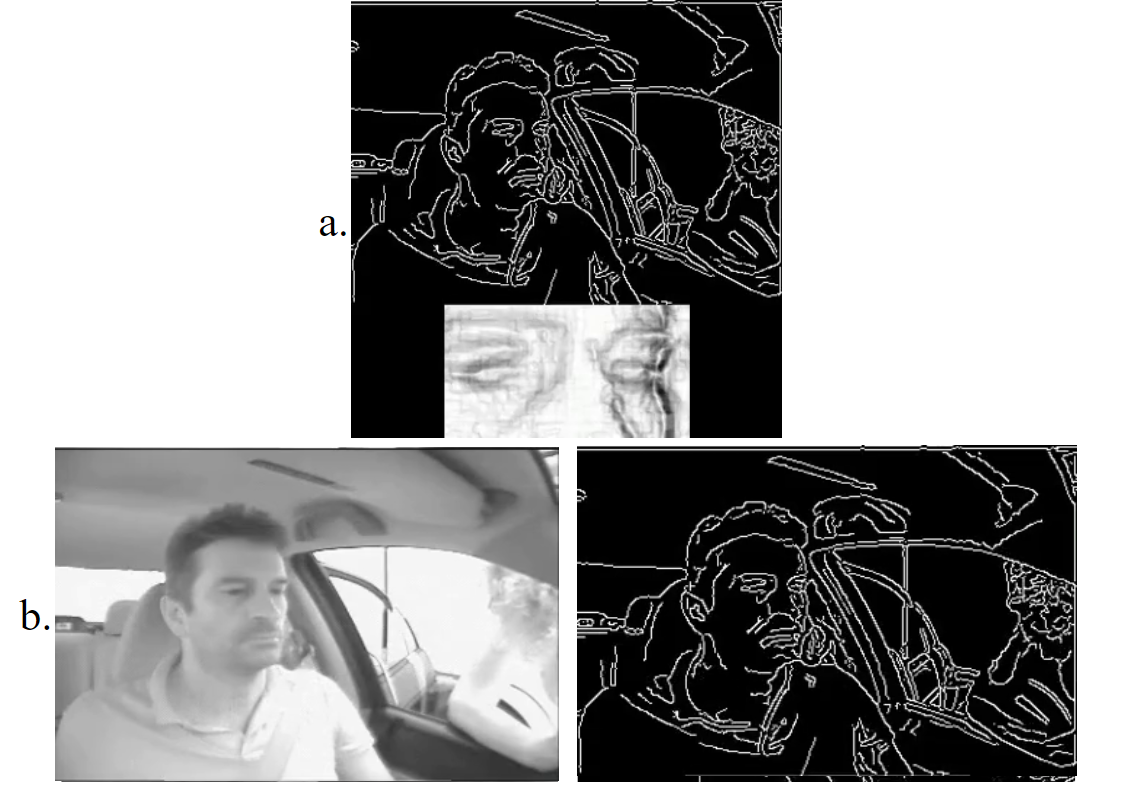}
      \caption{Example stimuli from the mask conditions. a. Canny+Eyezoom; b. ({\it left}) Unmasked, ({\it right}) Canny}
      \label{stim_conditions}
   \end{figure*}
The present work builds on previous work \cite{hooge2020evaluating}, with the goal of looking more carefully at the role the eyes play in facilitating face recognition in the context of identity masking. Simple filtering operations can preserve eye information, which is both valuable for gesture recognition, but may also inadvertently boost face recognition by people. Specifically,
in human perception experiments, the eye region of the face is known to support particularly good face recognition (e.g., \cite{royer2018greater}). In this study, we tested whether eye enhancement of an identity masked face would increase human face identification performance.  To that end, 
we created a set of stimuli in which the eye region was localized, expanded in size, and enhanced with a Scharr filter \cite{Jahne1999}. 
We compared face identification in three masking  conditions: 1.) unmasked driver videos,
2.) driver videos masked with the Canny method
\cite{Canny1986},
and 3.) a combination method that showed the Canny-masked video with an inset of the Scharr-enhanced eye region. See Figure \ref{stim_conditions} for an example of the stimulus conditions.  Note that we chose the Canny method filter for our masking algorithm, because it is relatively simple,  easy to implement, and is effective for both identity-masking and action preservation \cite{hooge2020evaluating}. 

In the first and second experiments, we focused on the effectiveness of identity masking. 
Videos were either shown unmasked (unaltered), masked solely with Canny, or masked with Canny and Canny+EyeZoom (see details, section \ref{conditions}).
The third experiment examined action preservation in these conditions.

\subsection{Study contributions}
\begin{itemize}
    \item  Masking the face of a driver in a video using a Canny filter effectively impairs face identification by comparison to an unmasked video.
    \item Enhancing and enlarging the eye region (Eyezoom of the face) and masking it with a Schaar filter does not alter the effectiveness of the Canny filter mask.
    \item Facial actions are preserved, in large part, when drivers' faces are masked with both the Canny and Canny + Eyezoom manipulations.
\end{itemize}

\section{Methods}

\subsection{Dataset}

Stimuli for the present experiment came from a set of driver videos in the Head Pose Validation (HPV) database.  The HPV dataset was created to emulate data from the  SHRP2-Naturalistic Driving Study (SHRP2-NDS) database \cite {Perez2016}, which is not easily available for research applications.  The 
SHRP2-NDS database  
is nearly unique in the range of imaging conditions encompassed in the data. It includes approximately 2 petabytes of video  from approximately $3,400$ drivers obtained over 1 to 2 years of observation. However, the dynamic video nature of the dataset provides for highly salient, personally identifiable, information about the drivers. The dataset is characterized by extreme illumination conditions (e.g., night-time shadowing, day-time bright spots, or illumination via transient headlights as a car turns). There is also the problem of quick driver movements (e.g., head turns and other actions which are very common in real-world driving). 

The HPV dataset used in the present study includes low-resolution videos of people actively driving a car or performing staged actions typical while driving, such as using a cellphone and putting on headwear or glasses. The video resolution is 356 × 240 pixels, with a frame rate of 14.98 frames per second. 
Each video segment was edited to 4s and masks were applied to the segments for direct comparison of mask effectiveness across conditions. Video length ranged from 1-4s depending on the type of action (looking left, looking right, looking down). The video segment lengths were identical for each identity across conditions.

\subsection{Conditions}\label{conditions}
The three masking conditions tested were implemented, as follows:

\begin{itemize}
\item {\bf unmasked} - drivers' faces were unaltered.
\item {\bf Canny mask} - drivers' faces were altered by applying a series of processes aimed at producing optimal edge detection, including the use of a Gaussian smoothing filter, a set of gradient-based edge detectors to enhance edges in the image, and then non-maximum suppression, threshold, and tracking to produce thin, refined edges.
\item {\bf Eyezoom condition}-- drivers' faces were first masked with the Canny process. Then the eyes were detected in the original image using the retinaface algorithm \cite {Deng2020}.  The original image was then expanded and masked with a Schaar filter, and the region around the eye detection was cropped.  Finally the Canny-masked face was presented in an inset showing the Schaar-filtered, zoomed eyes (see Fig. \ref{conditions}).  
\end{itemize}


    \begin{figure*}[h]
      \centering
      \includegraphics[width=.70\textwidth]{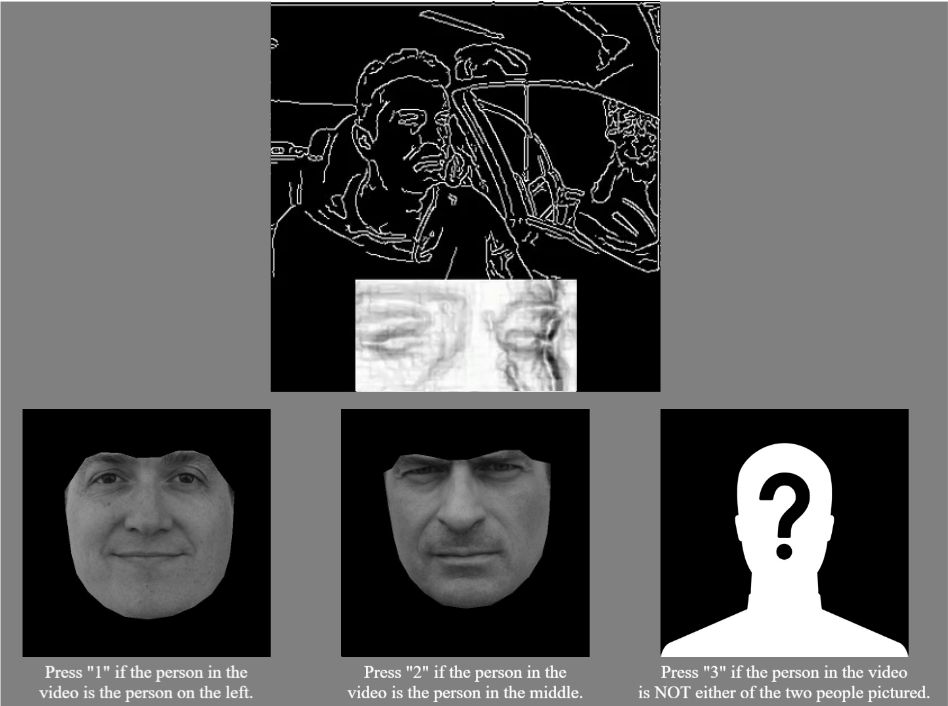}
      \caption{Example trial in Experiment 1. }
      \label{trial_E1}
   \end{figure*}

  \begin{figure}[h]
      \centering
      \includegraphics[width=.47\textwidth]{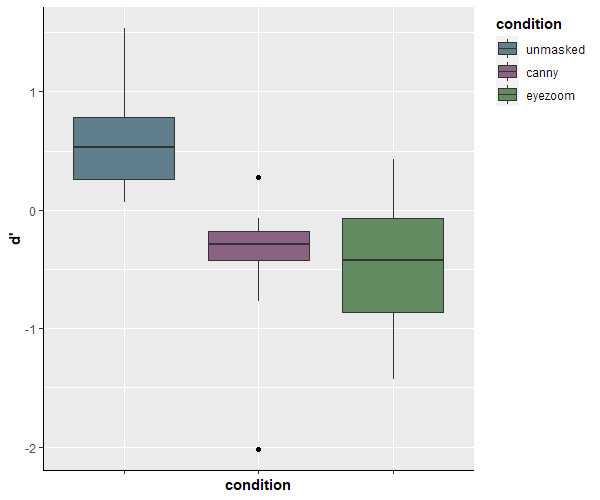}
      \caption{Experiment 1 accuracy, measured as {\it d'} across conditions. Results show that both masking algorithms were equally effective.}
      \label{E1_accuracy_figure}
   \end{figure}

 \begin{figure*}[h]
      \centering
      \includegraphics[width=.99\textwidth]{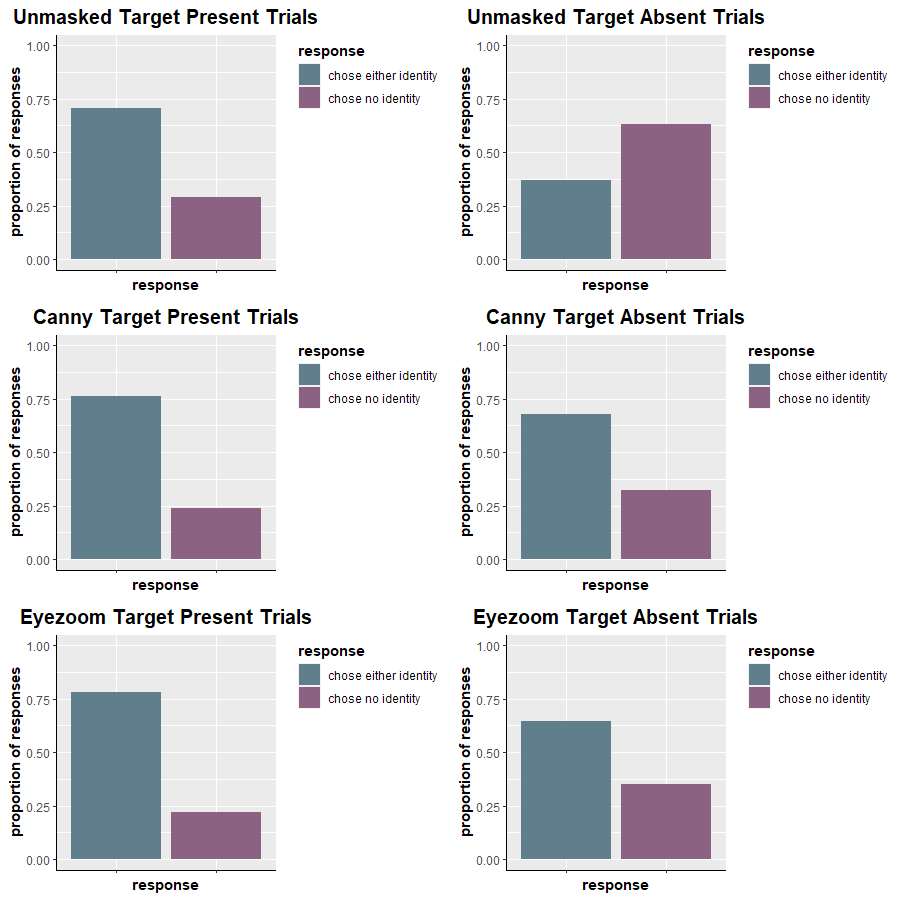}
      \caption{Proportion of responses by trial type in Experiment I.}
      \label{fig_resp_freq}
   \end{figure*}

\section{Experiment I: Effect of Eyezoom Masking Method}
In Experiment 1, we investigated the effectiveness of the Canny and Canny+Eyezoom filters at masking the identities of drivers in low-resolution videos.

\subsection{Participants}
A total of 30 (11 male, 18 female, 1 other) undergraduate student volunteers (ages 18-34) from the University of Texas at Dallas (UTD) participated in the study in exchange for research credit. All human experimental procedures were approved by UTD's Institutional Review Board.

\subsection{Procedure}
The experiment was composed of 72 trials 
in which 
a video stimulus was displayed in the top center of the screen.
The response options were presented below the video and showed
two faces and silhouette (see Figure \ref{trial_E1}).  
Participants were asked to select the face image that matched the driver in the video or to select the 
silhouette if neither of the two images matched the driver.
In target-present trials ($n=36$), one of the two faces 
matched the driver. In target-absent trials ($n=36$), neither 
of the two faces matched the driver.       
In all cases, the two face images presented as options showed
similar-looking identities from the dataset.  Each of the dataset's 36 identities was shown twice, once with the correct response being one of the target-present choices and once with the correct response being the target-absent choice.

 The video segments were shown in random order and looped until the subjects responded. Subjects were assigned randomly to one of the three masking conditions, with the unmasked condition serving as a control for general recognition success. Subjects were asked to determine whether the identity in the video matched one of the two identity images shown or if the identity was absent from the identity images shown.

\subsection{Outcome Measures}

\subsubsection{Accuracy}
Accuracy was assessed in two ways using a signal detection-type 
calculation based on 
{\it d'}.  This measure
depends on the proportion of hits $p(hit)$ and  false alarms $p(false \ alarms)$, as follows:

$$d' = z(p(hit)) - z(p(false \ alarms),$$

where the {\it z} refers to the z-score. 

In this experiment, hits were defined as target-present trials in which participants correctly recognized a driver as the matched-identity 
response choice. The design of the response options in the experiment offered two ways to compute false alarms.  Specifically,
false alarms can be defined as: a.)
target-present trials in which 
the participant choose the incorrect identity;
and/or b.) incorrect target-absent 
trials in which neither image showed the identity
(i.e., participants chose one of the face images, when neither was an identity match to the video). Because both options are consistent with the concept of a false alarm, in what follows, we included both types of false alarms (a and b) in the accuracy computation.


    \begin{figure*}[h]
      \centering
      \includegraphics[width=.47\textwidth]{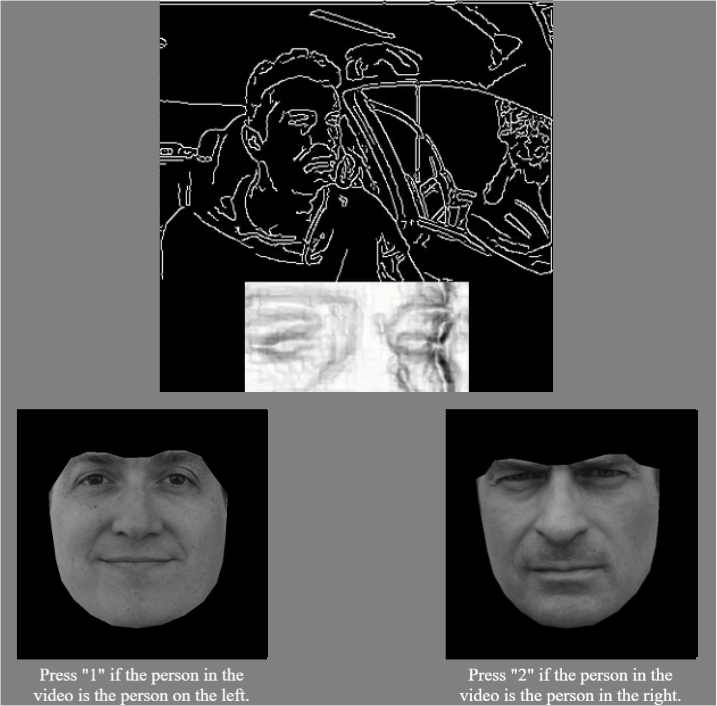}
      \caption{Example trial from Experiment II.}
      \label{E2_trial_fig}
   \end{figure*}
   
\subsection{Results}

\subsubsection{Accuracy} Figure \ref{E1_accuracy_figure} shows the average {\it d'} for each mask condition. These values indicate that faces in the unmasked condition were identified moderately well, but face recognition in both masked conditions was significantly impaired. 
The negative 
{\it d'} values for the masked conditions are unusual and suggest that participants used a 
systematically incorrect decision strategy, which we will investigate further 
in Section \ref{response_freq}.

   
A one-factor Analysis of Variance (ANOVA) was performed on  accuracy ({\it d'}), with mask condition as the independent variable. The resulting model yielded a main effect of mask condition on {\it d'}, $F(2, 27) = 11.03, p < .001$.  When comparing the conditions, {\it d'} accuracy was significantly higher in the unmasked condition than in the masked conditions, with no significant difference between the two masked conditions. This suggests that Canny and ORNL masking are not significantly less effective when used together than Canny masking alone. 
As is clear from the Fig. \ref{E1_accuracy_figure},
participant performance was more variable in the Eyezoom condition.

\subsubsection{Response Distribution}\label{response_freq}  
To further investigate the finding of negative {\it d}'s, we examined
 the proportion of responses allocated to each response type (face images chosen versus no identity chosen). The pattern of responses is shown for each mask type   in Figure \ref{fig_resp_freq}, with separate graphs  for  target-present (correct identity was available as a choice) and target-absent (correct identity was not available as a choice) trials.  For the unmasked condition, the graphs show a standard (relatively accurate) pattern of responses  as a function of whether 
the target was present or absent. The graphs for the masked conditions show inaccurate performance, but also suggest that participants did not systematically 
 choose the no-identity match when a match was present, but instead often chose the wrong face as the identity match.  
 
 We conclude tentatively that performance in the masked conditions was very poor indicating the effectiveness of the masks for preventing identification.  However, given the unusual performance in the masked condition (i.e., negative {\it d}'s), we retested the conditions with a design that eliminates the possibility of response bias.


\section{Experiment II: Effect of Eyezoom Masking Method with a Forced-Choice Task}

 In this experiment, we used 
 a two-alternative forced choice (2AFC) task
 to test masking effectiveness. In the  2AFC, two faces are presented as 
 response options.  In all cases, one of the two images will be the same identity as the person in the video.

\subsection{Participants}
A total of 30 (7 male, 22 female, 1 other) undergraduate student volunteers (ages 18-26) from UTD participated in the study in exchange for research credit.

\subsection{Procedure}
The experiment consisted of 72 trials. 
The video stimulus was displayed in the top center of the screen with the two face images beneath it. 
Participants were asked to determine which of the two face images matched the identity shown in the video.
To make the task challenging, the two faces presented had a similar appearance and were of the same race and gender.
An example trial is shown in Fig. \ref{E2_trial_fig}.

Each of the dataset’s 36 identities was shown twice, once with the correct response as the left-located option and once with the correct response as the right-located option. 
The video segments were shown in random order and looped until the subjects responded. Participants were assigned randomly to one of the three masking conditions, with the unmasked condition serving as a baseline condition for identification accuracy.


\subsection{Results}

\begin{figure}[h]
      \centering
      \includegraphics[width=.47\textwidth]{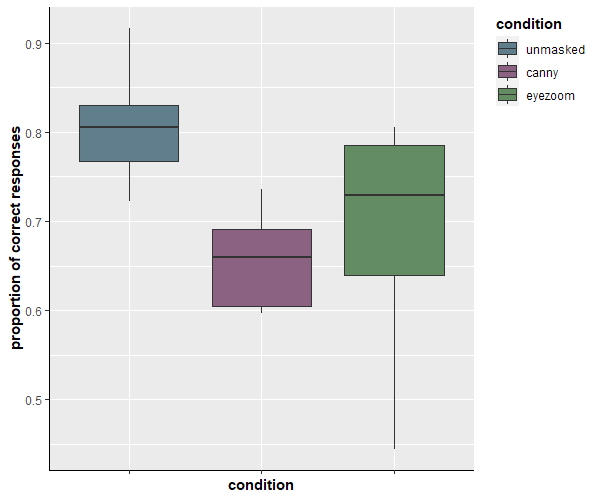}
      \caption{Experiment 2 - identification accuracy across conditions.}
      \label{E2_accuracy_figure}
   \end{figure}

Accuracy was assessed as the proportion of correct responses. Fig. \ref{E2_accuracy_figure} shows the proportion of correct responses for each mask condition. These values indicate that face recognition in the unmasked condition was more accurate than face recognition in the masked conditions.
A one-factor ANOVA was performed on the accuracy data (proportion of correct responses), with 
condition as the independent variable. The model yielded a main effect of mask condition on proportion of correct responses, $F(2, 27) = 9.68, p < .001$.  As in the first experiment,  
participants were more accurate in the unmasked condition than in the masked conditions, 
and performed comparably for the two masked conditions.


The results replicate the pattern of performance across conditions found for Experiment 1. As expected with a 2AFC task, performance was more accurate in all three conditions than it was in Experiment 1.  Notably, average identification was above chance in both masked conditions. Performance in the Eyezoom condition was more variable than performance in the Canny mask condition---replicating a similar finding in Experiment I.

We conclude that the masks strongly inhibit identification, but that when forced to guess between two images (with the assurance that one was an identity match), participants fared better than chance. Notwithstanding, applications of identity masking would rarely if ever be able to assure 
a human or machine system that one of two candidates was an identity match.
Our goal in applying this method here was to test examine the role of response bias in the unusual pattern of results found in Experiment 1.  
The present results suggest that these masking algorithms leave behind some residual identity information in the face that humans can exploit when the response decision is highly constrained.
 As noted, it is unlikely
 that that would be the case in any applied scenario, and so we conclude that these simple 
simple filtering procedures provide a 
reasonably high degree of identity protection.
Additionally, we conclude, albeit more tentatively,
that the eyezoom procedure does not improve identification significantly over the Canny procedure.


\section{Experiment III: Effect of Eyezoom Masking Method on Action Preservation}

The  effectiveness of the identity protection provided by these masks opens the question of whether this protection comes at the cost of
preserving information about facial actions. In this experiment, we examined whether the Canny and Canny+Eyezoom 
mask conditions impaired driver 
facial action perception.

\subsection{Participants}
A total of 30 (6 male, 23 female, 1 nonbinary) undergraduate student volunteers (ages 18-30) from UTD participated in the study in exchange for research credit.

\subsection{Procedure}
The experiment consisted of 100 trials, each with three response options: a.) driver looking to the left, 2.)  driver looking to the right, and 3.) driver looking down. Each of the 36 identities in the dataset appeared between two and three times, each with a different action (looking right, left, down). 
Prior to the start of the main experiment, a pilot test with only the unmasked condition was conducted to ensure that the actions were  identifiable in all videos. This test resulted in
the elimination of eight (of 108) videos segments
in which  actions were not recognizable
at sufficiently high levels of accuracy for inclusion in the action preservation study.

    \begin{figure*}[h]
      \centering
      \includegraphics[width=.47\textwidth]{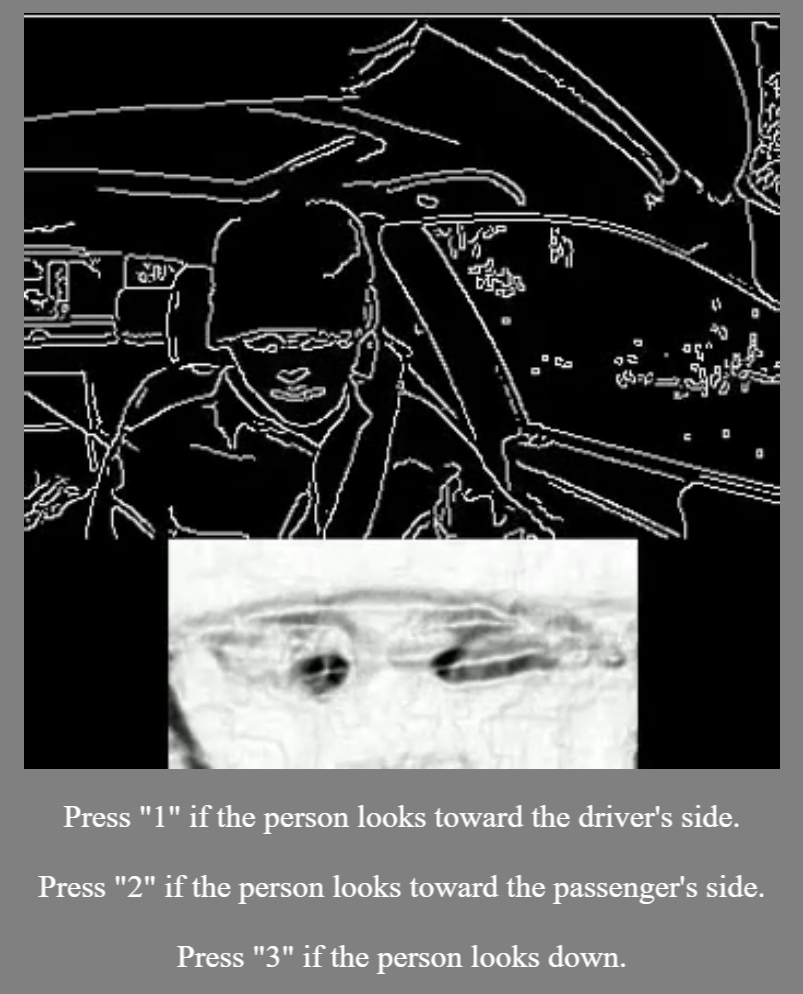}
      \caption{Example trial from Experiment III.}
      \label{E3_example_trial}
   \end{figure*}

The
participants were assigned randomly to one of three masking conditions with the unmasked condition providing a baseline action recognition accuracy and were asked to identify whether the driver was looking to the left, right, or down.
The video stimuli were shown in the upper center of the screen with three written options below.
See Fig. \ref{E3_example_trial} for an example trial.
The  clips were played in a random order and looped until the participant responded.

\subsection{Results}
\begin{figure}[h]
      \centering
      \includegraphics[width=.47\textwidth]{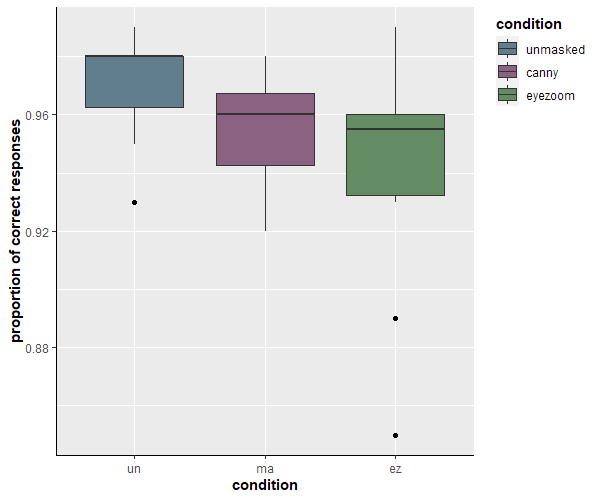}
      \caption{ANOVA of proportion of correct responses.}
      \label{E3_accuracy_figure}
   \end{figure}

The proportion of correct responses was used to assess accuracy. 
Fig. \ref{E3_accuracy_figure} shows the proportion of correct responses for each mask condition. These values indicate that action preservation was generally high, 
but also suggest a small advantage for action perception in the unmasked condition. 
A one-factor ANOVA, performed on the accuracy (proportion of correct responses) data, with the independent variable of condition, did not show a significant effect, but was generally consistent with this conclusion. The model yielded a marginal main effect of mask condition on proportion of correct responses, $F(2, 27) = 2.69, p = 0.086$.  
This suggests a very slight advantage for action perception without stimulus masking. 

In conclusion, although the results did not reach statistical significance, there is some indication that masking impaired action perception.


\section{Discussion} 
Our goal was to examine the effectiveness of simple Canny-filtering based masking methods, with and without eye enhancement, for interfering with face identification while preserving facial actions. In Experiment I, face recognition accuracy was diminished for both mask conditions, relative to the unmasked condition. There was no difference between the Canny mask alone and the mask with eye enhancement. 
In Experiment II, we replicated this result with a 2AFC procedure that controlled for response option bias, which may have been a factor in the findings of negative {\d'} values for both masking conditions.  In combination, both studies point to the relative effectiveness of the masks for interfering with identification. They also point to the conclusion that eye enhancement did not alter this effectiveness.  
Experiment III showed that facial actions were preserved to a similar degree with both masks, though there was a marginal advantage for action perception in the unmasked condition. 

In summary, these results indicate that Eyezoom masking does not significantly increase identification or alter facial action preservation.

\section*{Acknowledgment}
This work was supported through collaboration with Oak Ridge National Laboratory and the Federal Highway Administration under the Exploratory Advanced Research Program. The human experiment and analysis was subcontracted to the University of Texas at Dallas from Oak Ridge National Laboratory.

This manuscript has been authored in part by UT-Battelle, LLC, under contract DE-AC05-00OR22725 with the US Department of Energy (DOE). The US government retains and the publisher, by accepting the article for publication, acknowledges that the US government retains a nonexclusive, paid-up, irrevocable, worldwide license to publish or reproduce the published form of this manuscript, or allow others to do so, for US government purposes. DOE will provide public access to these results of federally sponsored research in accordance with the DOE Public Access Plan (http://energy.gov/downloads/doe-public-access-plan).

\bibliographystyle{IEEEtran}
\bibliography{AIM2_CNN_TBIOM}

\begin{thebibliography}{10}
\providecommand{\url}[1]{#1}
\csname url@samestyle\endcsname
\providecommand{\newblock}{\relax}
\providecommand{\bibinfo}[2]{#2}
\providecommand{\BIBentrySTDinterwordspacing}{\spaceskip=0pt\relax}
\providecommand{\BIBentryALTinterwordstretchfactor}{4}
\providecommand{\BIBentryALTinterwordspacing}{\spaceskip=\fontdimen2\font plus
\BIBentryALTinterwordstretchfactor\fontdimen3\font minus
  \fontdimen4\font\relax}
\providecommand{\BIBforeignlanguage}[2]{{%
\expandafter\ifx\csname l@#1\endcsname\relax
\typeout{** WARNING: IEEEtran.bst: No hyphenation pattern has been}%
\typeout{** loaded for the language `#1'. Using the pattern for}%
\typeout{** the default language instead.}%
\else
\language=\csname l@#1\endcsname
\fi
#2}}
\providecommand{\BIBdecl}{\relax}
\BIBdecl

\bibitem{hooge2020evaluating}
K.~D.~O. Hooge, A.~Baragchizadeh, T.~P. Karnowski, D.~S. Bolme, R.~Ferrell,
  P.~R. Jesudasen, C.~D. Castillo, and A.~J. O’toole, ``Evaluating automated
  face identity-masking methods with human perception and a deep convolutional
  neural network,'' \emph{ACM Transactions on Applied Perception (TAP)},
  vol.~18, no.~1, pp. 1--20, 2020.

\bibitem{Huang2012}
D.~Huang and F.~De~La~Torre, ``Facial action transfer with personalized
  bilinear regression,'' in \emph{Computer Vision--ECCV 2012}.\hskip 1em plus
  0.5em minus 0.4em\relax Springer, 2012, pp. 144--158.

\bibitem{xiong2013supervised}
X.~Xiong and F.~De~la Torre, ``Supervised descent method and its applications
  to face alignment,'' in \emph{Proceedings of the IEEE conference on computer
  vision and pattern recognition}, 2013, pp. 532--539.

\bibitem{DMask}
\BIBentryALTinterwordspacing
{Federal Highway Administration Active Project: Exploratory Advanced Research
  Program}, ``D{M}ask: A reliable identity masking system for driver safety
  video data.'' FHWA-PROJ-14-0054, 2014-2016. [Online]. Available:
  \url{https://highways.dot.gov/dmask-reliable-identity-masking-system-driver-safety-video-data}
\BIBentrySTDinterwordspacing

\bibitem{Canny1986}
J.~Canny, ``A computational approach to edge detection,'' \emph{IEEE
  Transactions on pattern analysis and machine intelligence}, no.~6, pp.
  679--698, 1986.

\bibitem{Jahne1999}
B.~J{\"a}hne, H.~Scharr, and S.~K{\"o}rkel, ``Principles of filter design,''
  \emph{Handbook of computer vision and applications}, vol.~2, pp. 125--151,
  1999.

\bibitem{Khojaste22}
M.~H. Khojaste, N.~M. Farid, and A.~Nickabadi, ``Gmfim: A generative
  mask-guided facial image manipulation model for privacy preservation,'' 2022.

\bibitem{royer2018greater}
J.~Royer, C.~Blais, I.~Charbonneau, K.~D{\'e}ry, J.~Tardif, B.~Duchaine,
  F.~Gosselin, and D.~Fiset, ``Greater reliance on the eye region predicts
  better face recognition ability,'' \emph{Cognition}, vol. 181, pp. 12--20,
  2018.

\bibitem{Perez2016}
M.~Perez, S.~Mclaughlin, T.~Kondo, J.~Antin, J.~Mcclafferty, S.~Lee, J.~Hankey,
  and T.~Dingus, ``Transportation safety meets big data: the shrp 2
  naturalistic driving database,'' \emph{Journal of the Society of Instrument
  and Control Engineers}, no. 55.5, pp. 415--421, 2016.

\bibitem{Deng2020}
J.~Deng, J.~Guo, E.~Ververas, I.~Kotsia, S.~Zafeiriou, and I.~FaceSoft,
  ``Retinaface: Single-shot multi-level face localization in the wild,''
  \emph{Proceedings of the IEEE/CVF conference on computer vision and pattern
  recognition}, 2020.

\end{thebibliography}

\end{document}